\documentclass{article} % For LaTeX2e
\usepackage{nips13submit_e,times}
\usepackage{hyperref}
\usepackage{url}
\usepackage{graphicx}
\usepackage{multirow}
\usepackage[numbers,sort&compress]{natbib}
\usepackage{mathtools}
\usepackage{amsmath}

\DeclarePairedDelimiter\floor{\lfloor}{\rfloor}

\title{Flood-Filling Networks}

\author{
Micha{\l} Januszewski \\
Google\\
\texttt{mjanusz@google.com} \\
\And
Jeremy Maitin-Shepard \\
Google \\
\texttt{jbms@google.com} \\
\And
Peter Li \\
Google \\
\texttt{phli@google.com} \\
\And
J{\"o}rgen Kornfeld \\
Max Planck Institute for Neurobiology \\
\texttt{kornfeld@neuro.mpg.de} \\
\And
Winfried Denk \\
Max Planck Institute for Neurobiology \\
\texttt{winfried.denk@neuro.mpg.de} \\
\And
Viren Jain \\
Google \\
\texttt{viren@google.com} \\
}

\nipsfinalcopy 

\begin{document}

\maketitle

\begin{abstract}

State-of-the-art image segmentation algorithms generally consist of at least two successive and distinct computations: a boundary detection process that uses local image information to classify image locations as boundaries between objects, followed by a pixel grouping step such as watershed or connected components that clusters pixels into segments. Prior work has varied the complexity and approach employed in these two steps, including the incorporation of multi-layer neural networks to perform  boundary prediction, and the use of global optimizations during pixel clustering. We propose a unified and end-to-end trainable machine learning approach, flood-filling networks, in which a recurrent 3d convolutional network directly produces individual segments from a raw image. The proposed approach robustly segments images with an unknown and variable number of objects as well as highly variable object sizes. We demonstrate the approach on a challenging 3d image segmentation task, connectomic reconstruction from volume electron microscopy data, on which flood-filling neural networks substantially improve accuracy over other state-of-the-art methods. The proposed approach can replace complex multi-step segmentation pipelines with a single neural network that is learned end-to-end. 

\end{abstract}

\section{Introduction}

A classic and prolific approach to image segmentation is to first transform an image into a graded representation called a ``boundary map'' that scores pixel locations as boundary or non-boundary, and then apply a flood-filling criterion such as connected components or watershed to generate segments from the boundary map. The simplest instantiation of this approach is to threshold pixel intensities and apply connected components. A more complex strategy, pursued by many state-of-the-art segmentation approaches, is to generate boundary maps using classifiers trained with supervised learning, and then apply customized watershed transformations to generate segments \cite{hwang2015pixel, arbelaez2011contour, kokkinos2015pushing, zlateski2015image}.

The incorporation of supervised learning into boundary detection has been crucial to achieving increasingly accurate segmentation results \cite{Martin:2004}. From the end-to-end machine learning point of view, however, a drawback of this pipelined approach is the fundamental disconnect between the learned part of the pipeline (pixel-wise boundary prediction) from the subsequent steps (connected components, watershed, etc) that produce the actual segments. The parameters of the boundary prediction function are optimized without consideration of the way in which individual predictions will ultimately be integrated by algorithm such as the watershed into a global interpretation of the image. Previous efforts to address this issue have focused on improvements to the cost function used during supervised learning of the boundary map classifier \cite{Turaga:2009, Jain:2010}, but have ultimately retained the final flood-filling step for producing object segments.  

In this paper, we propose a segmentation approach called ``flood-filling networks'' (FFNs) that forgoes explicit boundary detection and instead uses a single recurrent network to process raw image pixels directly into individual object masks. The primary contributions of this work are:
\begin{itemize}
\item A convolutional network architecture that introduces the notion of an ``object mask channel'' to both specify the target object and provide an explicit memory of segmentation state across recurrent iterations of the network.
\item A recurrent procedure for iterating the network inference dynamics over multiple overlapping fields of view in order to segment arbitrarily large objects that are initialized from a single seed pixel.
\item An experimental evaluation on a challenging connectomic dataset, demonstrating how a single flood-filling network can outperform a highly optimized and state-of-the-art pipeline consisting of three successive algorithmic steps: boundary detection by 3d convolutional neural networks, affinity graph watershed segmentation, and pairwise object agglomeration. 
\end{itemize}

\section{Related work}
A variety of papers have recently introduced ``instance'' segmentation methods that utilize recurrent or feed-forward networks to directly produce segmentation masks from an image. For example, \citet{romera2015recurrent} propose a recurrent architecture that at each iteration of the network dynamics produces a complete segmentation mask for a single object in the image. \citet{pinheiro2016learning} propose ``SharpMask'', a feed-forward architecture in which a convolutional network first generates coarse-resolution object masks, and then upsampling modules refine these masks at the original image resolution (resulting in an overall architecture similar to U-Net as developed by \citet{ronneberger2015u}). Our approach builds on this prior work, but is distinct in several respects:
\begin{itemize}
\item Unlike \citet{romera2015recurrent}, FFNs utilize multiple iterations of the network to delineate a single object. Therefore objects within the mask can grow or shrink across iterations depending on the network output, and the maximal size of a segmented object is unrestricted relative to the network architecture (this property is important for the application of connectomics, where a single object may span millions of voxels).
\item Unlike \citet{pinheiro2016learning}, the entire architecture is trained end-to-end. 
\item Unlike \citet{pinheiro2016learning}, FFNs utilize a concept of an object mask channel to specify the object being segmented, rather than treating the central pixel in the field of view as special. 
\end{itemize}

\section{Flood-Filling Networks}

\subsection{Architecture}
A flood-filling network takes a 3d subvolume of data as input and produces an object mask probability map. In our implementation, the output probability map is of the same size as the input subvolume. The input subvolume contains at least two channels: one providing the raw image intensities, and another providing the local state of the object mask in the form of a probability map. The input mask in the second channel is incomplete in most inference iterations of the network, and the FFN is trained to extend it within its field of view (FoV). At the beginning of inference, in absence of any prior information about object shape, the incomplete object mask can consist of a single active voxel. The network then performs single-object segmentation for the object covered by that voxel.

While a single run of the network necessarily provides an object probability map limited to the FoV of the network, objects of arbitrary size can be created by letting the network ``write'' the probability map to a virtual canvas, and repeatedly performing network inference while moving the center of the FoV, and providing the current local state of the canvas in the object mask channel of the input.

The network consists of a deep stack of 3d convolutions with ReLU nonlinearities. Since the spatial size of the input and output of the FFN are identical, all convolutions use the SAME mode, and no pooling is used anywhere in the network -- an architecture that is nearly identical to the one used in the first work that had applied convolutional networks to connectomic reconstruction \cite{Jain:2007} and that is similar to the more recent PixelCNN model \cite{oord2016pixel}. The SAME mode for a convolutional layer with the kernel size $k_x \times k_y \times k_z$ is implemented by padding the input with $\floor{k_x / 2}$, $\floor{k_y / 2}$, $\floor{k_z / 2}$ zeros on each side, respectively in the $x$, $y$, and $z$ dimensions.

The spatial size of the network FoV is set to $33 \times 33 \times 17$ voxels, corresponding roughly to a cube in physical space due to the anisotropy of the underlying image data. All convolutions use $3 \times 3 \times 3$ filters. The depth of the network is set so that information from the whole input subvolume contributes to the value of the central voxel of the output, which implies at least 17 convolutional layers. We tried varying the depth, and determined empirically that a lower depth causes the network to perform worse at evaluation time, and additional layers beyond the minimum 17 did not yield further improvements.

\begin{figure}
    \centering
    \includegraphics[scale=0.4]{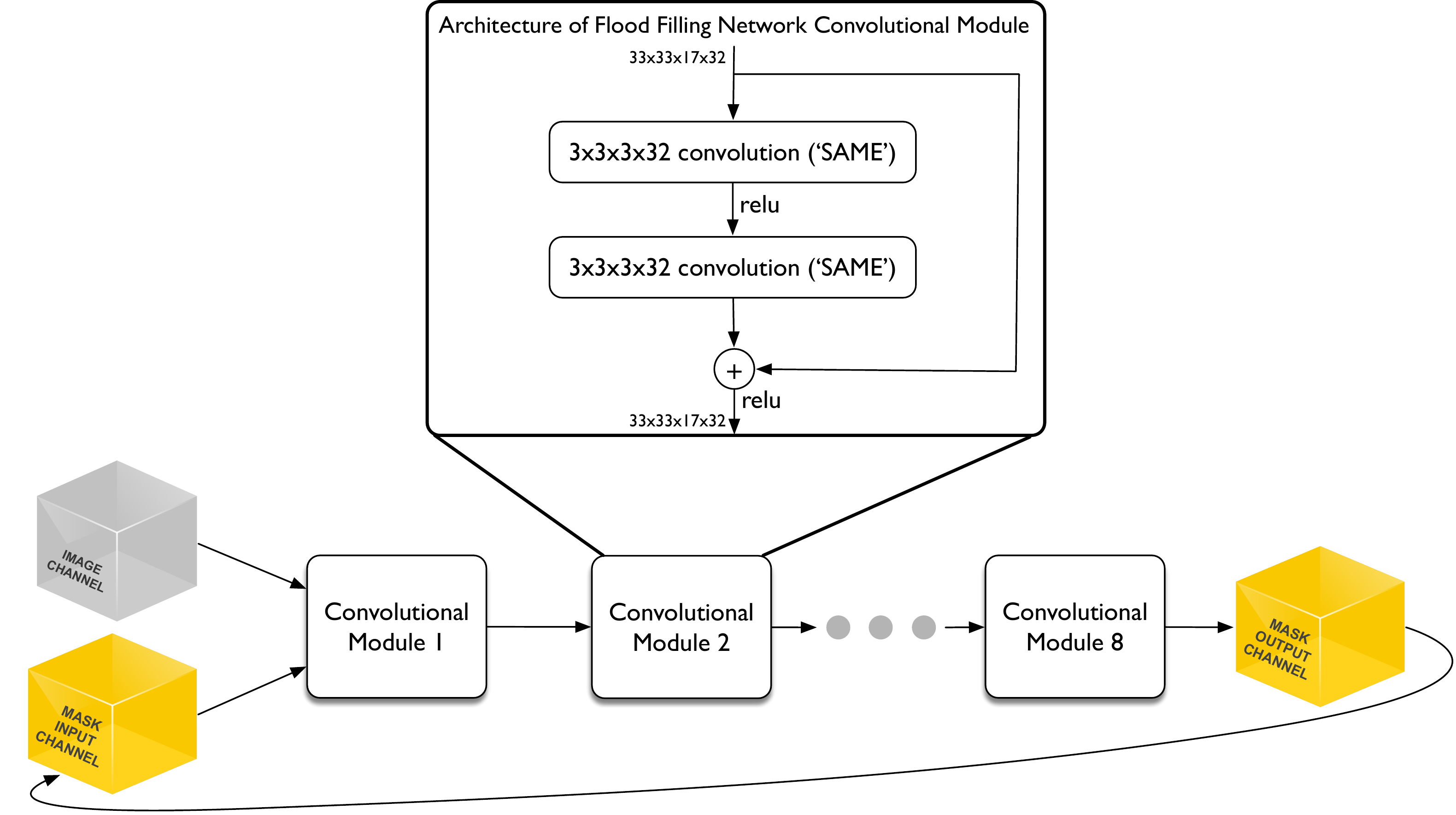}
    \caption{Architecture of a flood-filling network.}
    \label{fig:architecture}
\end{figure}

Internally, the network is organized into modules of two convolutional layers, with skip connections between them  (see Figure~\ref{fig:architecture}), similar to the residual units of  \citet{he2015deep}. We experimented with several variants of skip connections, and settled on an architecture equivalent to ``full pre-activation'' of \citet{he2016identity}, which resulted in best network performance.

\subsection{Learning and Inference}
The network is trained to extend an input object mask into the full extent of the object within the network's field of view. A special case is the first iteration of the network, in which a single pixel is encoded in the center of the input object mask to specify the particular object the network should be extending. A training example consists of an object fragment contained within a $49 \times 49 \times 25$ subvolume, with the object of interest overlapping the center voxel of the subvolume. The objective of the network is to predict the voxel mask of the object, with a value of $0.95$ indicating that a given voxel belongs to the object of interest, and a value of $0.05$ that it does not. The values $0.05$ and $0.95$ are chosen as soft-target equivalents of $0$ and $1$, with the $0.05$ softness parameter ensuring that the internal activations of the network are not pushed towards $\pm \infty$.

In the first step of training, the FoV of the network is set to the center of the $49 \times 49 \times 25$ subvolume of the current training example, and the center voxel of the input object mask channel is set to the target value of $0.95$. The remaining voxels are set to $0.05$. The network is then run in forward mode, resulting in a partial predicted mask covering the current FoV. This mask is saved within a ``canvas'' covering the whole training subvolume and the FoV of the network is moved to a new location.

\begin{figure}
    \centering
    \includegraphics[scale=0.45]{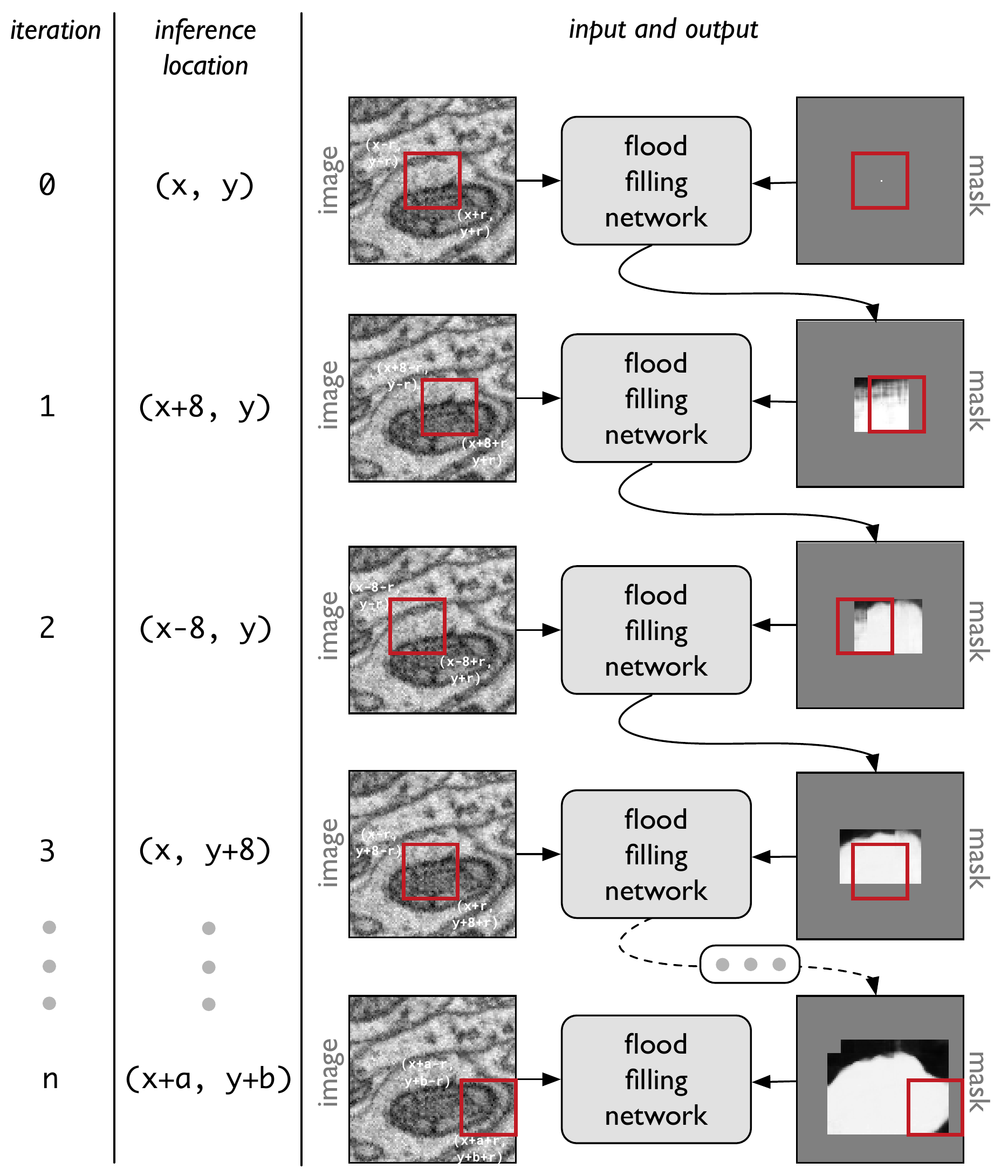}
    \caption{Schematic of multiple-field-of-view inference of a flood-filling network.}
    \label{fig:inference}
\end{figure}

Every FFN has an associated $\vec{\Delta}$ vector determining the step size to take in every direction when moving the FoV, as well as a threshold value $t_{\mathrm{move}}$ which needs to be reached or exceeded in the current object mask at the voxel that is to become the new center of the FoV. For the network used in this work, $\vec{\Delta} = (\Delta_x, \Delta_y, \Delta_x) = (8, 8, 4)$ and $t_{\mathrm{move}} = 0.9$. For brevity, we will refer to the position of the center of the FoV of the network simply as the \emph{position} of the FFN. If we denote the current position $(x_0, y_0, z_0)$, potential new positions are searched by examining the current state of the mask at the 6 planes $x = x_0 \pm \Delta_x$, $y = y_0 \pm \Delta_y$ and $z = z_0 \pm \Delta_z$ further restricted to $[x_0 - \Delta_x \leq x \leq x_0 + \Delta_x] \times [y_0 - \Delta_y \leq y \leq y_0 + \Delta_y] \times [z_0 - \Delta_z \leq z \leq z_0 + \Delta_z]$. For every such plane the object mask voxel with the highest value $v$ is found, and if $v \geq t_{\mathrm{move}}$ the location of that voxel is added to a list of new positions for the FFN. Once all 6 planes are examined, the list is sorted in descending order of activation of the corresponding mask voxel and appended to a queue of new positions. Inference proceeds by pulling new positions from the queue and moving the FFN to them. After every such move, the search for new locations relative to the current position of the network is repeated. Inference ends when the queue is empty. To prevent the network from getting stuck in a loop, a set of already visited locations $V$ is maintained. For efficiency, the locations are stored at a reduced resolution as $(\floor{x_0 / \Delta_x} ,\floor{y_0 / \Delta_y}, \floor{z_0 / \Delta_z})$. The network is forbidden from visiting locations that are already in $V$. A simplified version of this procedure is shown in Figure~\ref{fig:inference}.

During training, the inference procedure described above is used, but network moves are restricted so that the FoV never leaves the training example $49 \times 49 \times 25$ subvolume. A voxelwise log-loss is computed against the target object mask after every FoV move and the network weights are updated by stochastic gradient descent. If $v_i$ is the predicted mask value for the $i$-th voxel of the current FoV at the output of the network, and $m_i$ is the target value for that $i$-th voxel, the loss optimized during training is:
\[
    \sum_{i=0}^{33 \times 33 \times 17} -\log(v_i)  m_i -\log(1 - v_i) (1 - m_i).
\]

Inference proceeds similarly to training, but the network is no longer restricted to a small subvolume, and we explicitly bias the network against merge errors by making it impossible to reverse split decisions. An object mask voxel $v$ can be updated by the network multiple times during the inference procedure. If we denote the $t$-th value of that voxel $v_t$, with $v_0$ standing for initial state of the voxel before the first step of inference, and if $v'_t$ is the value predicted by the network based on the previous state $v_{t-1}$, the split bias is implemented as:
\[
    v_t = \begin{cases}
        v_{t-1} & \text{if $v_t > v_{t-1}$, $v_{t-1} < 0.5$, and $t > 1$,} \\
        v'_t & \text{otherwise.}
    \end{cases}
\]

Inference also requires a way to select the location of the initial seed points from which the network will attempt to grow object masks. There are various ways this could be done -- e.g. by choosing points at random or on a grid. In our experiments, we found it beneficial to use a simple heuristic that attempts to find points maximally distant from suspected object boundaries. To find these points, we process the raw image data with a 3d Sobel filter, compute the distance transform of the result and choose local maxima of the distance transform as the seed points for the FFN.

To convert the object probability map generated by the network into a hard segmentation, we simply threshold the probability map at $t_{\mathrm{move}}$, and assign IDs to the resulting segments based on the seed number used to run the FFN inference.

\subsection{Implementation}
We implemented the FFN model in TensorFlow \cite{abadi2016tensorflow}, and trained it with asynchronous gradient descent in a distributed setting using 32 NVIDIA K40 GPUs, exploiting data parallelism. The learning rate was set to $0.001$, and the batch size to 4 (further increases did not improve performance on the GPU). 3d convolutions were executed by CuDNN R4 and used single-precision floating point numbers.

Training examples were extracted at random as $49 \times 49 \times 25$ subvolumes from the 33 densely annotated training volumes. The size of the subvolume was chosen to allow one step of FoV movement in every direction, as described above. The ground truth segmentation within every subvolume was binarized by setting voxels belonging to the same object as that of the central voxel of the subvolume to the value $0.95$, and the rest of the voxels to $0.05$, thus providing soft targets for the desired object mask probability map. For every training example, the fraction of active mask voxels $f_a$ was calculated. The training examples were then divided into 17 classes, such that an example was assigned to class $i$ if $t_{i-1} \leq f_a < t_i$, and $t_i = (
0.0, 0.01, 0.02, 0.03, 0.04, 0.05, 0.06, 0.075, 0.1, 0.2, 0.3, 0.4, 0.5, 0.6, 0.7, 0.8, 0.9, 1)$. The examples were then resampled to achieve equal distribution of these classes. The dense sampling of low-$f_a$ examples was chosen to make the network sensitive to thin processes such as axons and dendritic spine necks.

To monitor progress of the training procedure, weight checkpoints were saved every 1 h, and for each checkpoint an inference process covering a densely skeletonized subvolume (see Section~\ref{sec:dataset}) was automatically performed. The segmentation results were then evaluated using skeleton metrics (see Section \ref{sec:metrics}). Training was terminated when the metrics stopped improving.

\section{Connectomic Reconstruction Experiments}
We evaluated flood-filling networks on the task of connectomic reconstruction of brain tissue from high resolution volume EM data. The ability to map wiring diagrams of neural circuits (``connectomes'') remains limited due to technical difficulties involved in nanometer-resolution imaging and the currently mainly manual neurite reconstruction process required for analysis. In particular, the task of segmenting densely interwoven neurons and neurites in teravoxel-scale 3d images remains especially problematic for algorithmic approaches [24], where even state-of-the-art deep learning pipelines have thus far yielded results insufficiently accurate for purely automated analysis~\cite{maitin2015combinatorial}. Thus, 3d segmentation of volume EM datasets is an important area for continued progress of machine perception capabilities.
The ground truth data in this field is also well defined and unambiguous in interpretation, enabling accurate comparisons for different image segmentation algorithms

\subsection{Dataset}
\label{sec:dataset}
Neuropil of the zebra finch songbird (\textit{Taeniopygia guttata}) Area X was imaged using Serial Block-face Scanning Electron Microscopy (SBEM) at $10 \times 10 \times 20$~nm resolution [11]. Out of the complete $\sim$500-gigavoxel dataset, 29 $150^3$ voxel and 4 $256\times256\times128$ voxel spatially-separated volumes were manually annotated with dense neurite segmentations. A separate
$520 \times 520 \times 256$ volume was densely skeletonized using Knossos (\url{http://knossostool.org/}) and used here as a testing set.

Within this volume, 221 disconnected fragments were skeletonized with a total of 5234 edges, corresponding to a  path length of $\sim1$~mm. The subvolume covers an
area of dense neuropil. A small cell body slice, as well as 3 glial fragments are also included.

\subsection{Baseline Segmentation Approach}
In order to evaluate flood-filling networks, we established a competitive baseline that reflects state-of-the-art practice in connectomic segmentation: deep learning based boundary prediction \cite{Jain:2007, ciresan2012deep}, watershed over-segmentation \cite{zlateski2015image}, and supervoxel agglomeration \cite{maitin2015combinatorial, nunez2014graph}. Specifically, we trained a 3d convolutional neural network (precise architecture details are provided in Table 2 of \cite{maitin2015combinatorial}) to predict, based on a $35 \times 35 \times 9$ voxel image context region, whether the center voxel is part of the same neurite as the adjacent voxel in each of the $x$, $y$, and $z$ directions, as in prior work \cite{Turaga:2010uq}. We optimized the parameters of the neural network using stochastic gradient descent with log loss. 
Using the affinities generated by the 3d convolutional network, we applied a watershed algorithm [14] with the goal of achieving an (approximate) oversegmentation using parameters $T_l, T_h = \{0.945, 0.95\}$, $T_e = 0.5$, and $T_s = \{100, 1000\}$~voxels. Finally, we agglomerated the segments produced by watershed using GALA \cite{nunez2014graph} and CELIS \cite{maitin2015combinatorial}, which were trained using the 33 cubes of volumetric ground truth. 

\subsection{Skeleton Evaluation Metrics}
\label{sec:metrics}

Although segmentation quality is often evaluated with respect to ground truth pixel-wise contours or object masks \cite{Martin:2004}, creating such ground truth for large-scale connectomic datasets that span billions or trillions of voxels is highly laborious. A more efficient way to generate ground truth representations of large-scale neuron topology is to ``skeletonize'' neurons into a collection of points that typically constitute an undirected tree \cite{helmstaedter2011high}.

In order to evaluate the accuracy of the proposed segmentation algorithm, a $560 \times 560 \times 250$ voxel subvolume from the dataset was densely skeletonized by humans (i.e., every object in the subvolume was annotated with a skeleton). We then computed the accuracy of machine-generated volumetric segmentations of the same subvolume by comparing to the ground truth skeletons using the following criteria. 

\subsubsection{Edge Accuracy}
\label{sec:edgeacc}

We would like a single number that represents the fraction of the skeleton topology that is correctly reconstructed. We assume:
\begin{itemize}
\item a ground-truth skeleton $S_i$ consists of edges $\left\{ e_1, e_2, \ldots, e_{|S|} \right\}$,
\item an edge $e$ is defined by two 3d node coordinates $A(e)$ and $B(e)$,
\item $S(e)$ denotes the ID of the ground-truth skeleton containing edge $e$,
\item a predicted segmentation $R$ to be evaluated, where $R(p)$ returns the value (object ID) at point $p$. $R(e)$ is a shorthand for ``either $R(A(e))$ or $R(B(e))$''.
\end{itemize}

\begin{figure}
    \centering
    \includegraphics[width=0.15\textwidth]{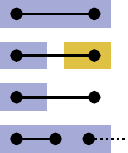}
    \caption{Different types of edges that are involved in the skeleton edge accuracy metric; colors correspond to segment IDs. From top to bottom: correct edge (both nodes have the same ID), split edge (nodes assigned to different segments), omitted edge (one or two nodes do not have an associated segment ID), merged edge (node assigned to a segment that covers more than one skeleton).}
    \label{fig:edges}
\end{figure}

An edge $e$ is defined as correctly reconstructed if the nodes of the edge are assigned to the same object in $R$ and no nodes from different skeletons reconstruct to that same object in $R$. More formally, we classify every edge $e$ into one of four categories (see Figure~\ref{fig:edges}):
\begin{itemize}
 \item \emph{omitted} if $R(e) = 0$,
 \item \emph{split} if $R(A(e)) \neq R(B(e))$,
 \item \emph{merged} if there exists $e_m$ such that $R(e) = R(e_m)$ but $S(e) \neq S(e_m)$,
 \item \emph{correct} if none of the above is true.
\end{itemize}

The \emph{edge accuracy} is therefore defined as the percentage of correctly reconstructed edges over all the ground truth skeletons. Incorrect edges can be further subdivided into the percentage of edges which have a merge, split, or omitted error. To compensate for the variability of the human-generated skeleton node positions within neural processes, we allow omitted edges to be treated as correct if one of the edge nodes is of degree one (i.e., the edge is a leaf of the skeleton) or if the omitted edges form a subgraph that is connected to two or more nodes associated with the same non-zero segment ID (i.e., the subgraph does not break continuity of the object within which it is contained). The latter adjustment also allows us to ignore cases where an internal organelle got excluded from the segmentation and marked as background without affecting the continuity of the neurite containing the said organelle.

\subsubsection{Comparison to SegEM split/merger counts}
\citet{berning2015segem} define the number of splits and mergers in a predicted dense segmentation with respect to a set of ground truth skeletons as follows:
\begin{enumerate}
\item An individual skeleton is said to correspond to a predicted segment if at least $k$ nodes within the skeleton are contained within the predicted segment, for $k = 1$ or $2$.  The stated reasoning for picking $k > 1$ is for greater robustness to imprecise placement of skeleton nodes.
\item For each skeleton, the number of splits is determined as max(0, number of corresponding predicted segments - 1).
\item For each predicted segment, the number of mergers is determined as max(0, number of corresponding skeletons - 1).
\end{enumerate}
\citeauthor{berning2015segem}'s definitions result in a metric with different properties from the one proposed above. Specifically, in their metric merging two predicted segments may decrease the number of merge errors; in the limit case of a single predicted component encompassing the entire volume, the number of merge errors is (1 - number of skeletons).

We also note that our edge accuracy metric heavily penalizes merge errors since \emph{all} edges covering the merged predicted segments are considered erroneous. We argue that this a desired property of the metric, reflecting the significantly higher effort required to fix such errors manually during proofreading of the automated results. 

\section{Results}

\begin{table}
	\centering

\begin{tabular}{ l | c c c c c}
\hline 
\multirow{3}{*}{Segmentation} & Edge & Merged & Split & Omitted & Omitted \\
& accuracy & edges& edges  & edges & edges \\
  & [\%]  & [\%]  & [\%]  & (adjusted) [\%] & (raw) [\%] \\
\hline 
CNN + Watershed & 	    	87.7 &		1.0 &		10.6 &		0.7 &      1.1 \\
CNN + Watershed + GALA & 	96.3 &		1.7 &		1.4 &		0.6 &      1.1 \\
CNN + Watershed + CELIS &	93.2 &		5.4 &		0.7 & 		0.7 &      1.1 \\
Flood-Filling Network & 	98.5 &		0.0 &		0.7 &		0.8 &      2.4 \\
\hline 
\end{tabular}
\caption{Edge accuracy results as measured on a $520 \times 520 \times 256$ test volume (5234 total edges in the ground truth skeletons). Merged, split, and omitted percentages correspond to different classes of errors (lower is better). The ``adjusted'' and ``raw'' omitted edge fractions correspond to omitted edge counts respectively with and without allowing some such edges to be treated as correct, as explained in Section~\ref{sec:edgeacc}.}
\label{tab:results}
\end{table}

\begin{figure}
    \centering
    %\begin{subfigure}
    %    \centering
        \includegraphics[width=0.49\textwidth]{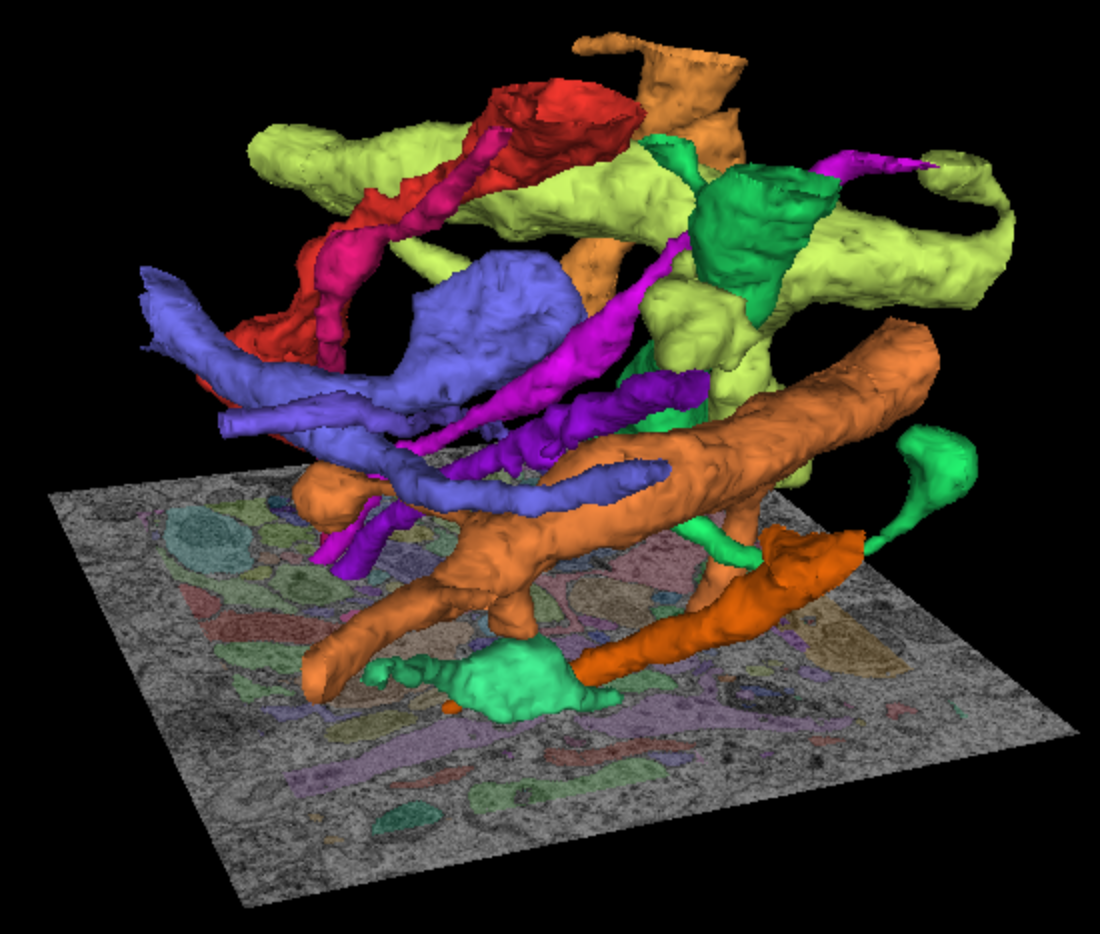} 
    %\end{subfigure}
    \hfill
    %\begin{subfigure}
    %    \centering
        \includegraphics[width=0.49\textwidth]{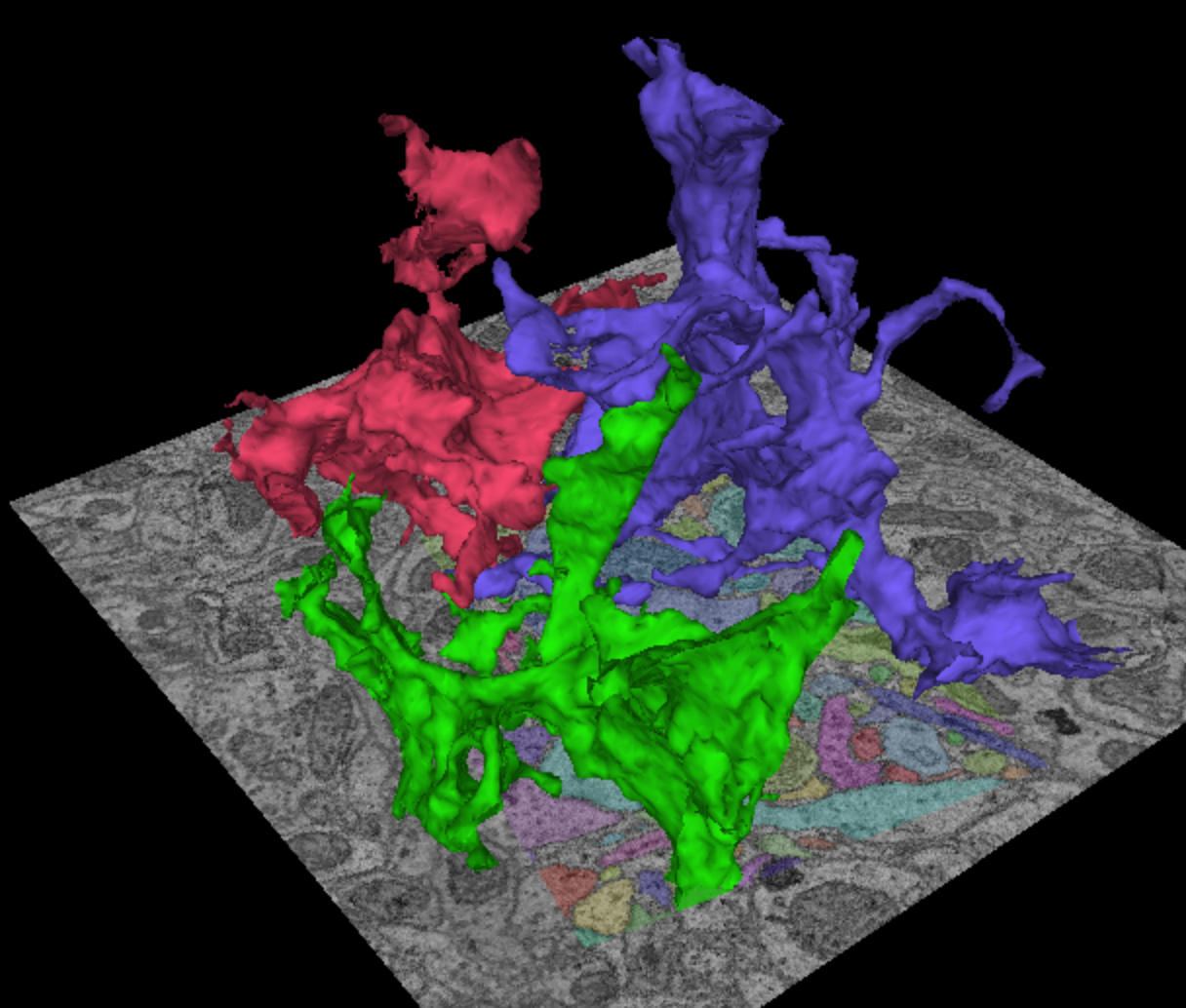} 
    %\end{subfigure}
    \caption{Visualization of selected processes within the FFN reconstruction of the test subvolume.
    The shown EM data corresponds to a $6.6 \times 6.6\,\mathrm{\mu m}$ tissue fragment.
    Left panel: a mix of dendrites and axons. Less than 6\% of all
        fragments contained within the subvolume is shown. Right panel: largest reconstructed fragments from 3 glia in the subvolume.}
    \label{fig:render}
\end{figure}

We evaluated all automated segmentations using the metrics described in the previous section, and report the results in Table~\ref{tab:results}. Only the best result for a given method, as measured by edge accuracy, is reported. The best result corresponds to:
\begin{itemize}
\item for CNN + Watershed: optimal affinity graph watershed settings $T_l, T_h = 0.945$, $T_e = 0.5$, and $T_s = 1000$,
\item for GALA and CELIS: the optimal agglomeration threshold score,
\item for FFN: the optimal checkpoint of the neural network parameters. Note that we currently observe a significant amount of instability in the performance of the FFN from one checkpoint to the next; addressing this instability by improvements to the training procedure is a high priority for future work. 
\end{itemize}
We note that no additional FFN parameters (such as  $t_{\mathrm{move}}$) were optimized. A visualization of the FFN segmentation is shown in Figure~\ref{fig:render}.

In comparison with all the other methods, the FFN segmentation stands out as the one with the highest edge accuracy, while also being the only one able to completely avoid merge errors. 
We have observed that when reconstructing volumes much larger than the $520 \times 520 \times 256$ subvolume evaluated here, the superior ability of FFNs to avoid merge errors accumulates to a dramatic practical advantage over alternative methods. 

The higher raw omitted edge count for the FFN segmentation shown in Table~\ref{tab:results} is caused by the more compact object masks generated by the network (in comparison to watershed), as well as a tendency of the network to sometimes miss small internal areas of larger processes without breaking their overall continuity. The latter case is usually associated with the presence of an organelle, such as a mitochondrion. As these issues are irrelevant for the purpose of building accurate connectomes, and could be relatively easily fixed in post-processing if needed, we used the adjusted omitted edge fraction in the calculation of overall edge accuracy.

\section{Discussion}
Flood-filling networks are both simple and effective, but  also more computationally expensive than traditional approaches. 
\subsection{Simplicity}
FFNs are simple in that the architecture of the network is a series of convolutions with skip connections. There are no pooling, subsampling, upsampling, or other types of layers. A nearly identical architecture was one of the first approaches used for applying machine learning to connectomic reconstruction \cite{Jain:2007, Turaga:2010uq}, and more recent approaches for segmentation in other domains have also begun exploiting so called ``fully-convolutional'' architectures \cite{oord2016pixel}. 

The key difference from prior work in \cite{Jain:2007, Turaga:2010uq} is the learning and inference scheme employed by FFNs. Instead of generating a dense pixel-wise affinity graph or boundary map for all objects in the field of view associated with a forward pass of the network, each forward pass of the FFN produces only the object mask associated with a \textit{single} object. Multi-object segmentations are sequentially constructed by multiple iterations of the network on overlapping fields of view. This approach naturally addresses a discrepancy between the local, per-pixel cost functions used to train segmentation networks and the potentially non-local effects of errors. Addressing this discrepancy has previously required highly sophisticated manipulations to traditional neural network training schemes \cite{Turaga:2009, Jain:2010}. In contrast, the single-object loss and gradients computed in FFN training naturally represent the large errors that might accumulate from a critical under-segmentation  or over-segmentation error. The network can in fact even learn to correct its own mistakes through multiple iterations of the inference dynamics (for example, as it processes more of some critically relevant image context). 

\subsection{Efficacy}
The results presented in Table~\ref{tab:results} demonstrate that FFNs can substantially exceed the accuracy of  established connectomic segmentation techniques that also rely on deep learning as a core component. A particularly interesting aspect of the results is that a single FFN network can outperform a multi-step pipeline of convolutional network boundary prediction, watershed over-segmentation, and machine learning based supervoxel agglomeration with either GALA or CELIS. This pipeline consolidation drastically reduces engineering complexity and hyperparameter tuning needed to deploy such algorithms in practical reconstruction workflows. Moreover, the zero percent merge rate demonstrated by FFNs is especially promising, as merge errors tend to cause the most serious and difficult to handle errors in human-assisted semi-automated connectomic reconstruction techniques. 

\subsection{Computational Cost}
One downside of the current implementation of FFNs is their significantly increased computational cost caused by the depth of the network, and the single object segmentation scheme.
Specifically, for the densely skeletonized subvolume used in this work the cost of the 3d affinity graph CNN inference is approximately 0.14~PFLOP (applying GALA and CELIS introduces additional computations that are a similar order of magnitude or less),
while the total cost of the FFN segmentation is 4.6 PFLOP.
While inference cost is still proportional to the size of the volume to be segmented, FFNs introduce an additional factor proportional
to the number of distinct objects present within the FoV of the network. This factor obviously varies depending on the precise location in
the neuropil, but in practice 2-3 objects can be expected to be present on average within the FoV for the dataset and network architecture used in this work.
We note also that every voxel within a segmented object is processed multiple times as the FoV movement step size is small enough that
subsquent FoVs overlap to a significant degree (42 - 75\%).

\subsection{Future Directions}

We have identified three major directions for further development of FFNs aimed at: 1) reducing the computational cost, 2) improving segmentation precision, 3) removal
of heuristic elements within the training and inference procedures.

Within the first group, we propose architecture changes which reduce the number of necessary floating point operations by incorporation of a bottleneck layer
similar to U-net~\cite{ronneberger2015u}, use of strided convolutions for faster spatial information propagation within the network, larger FoV step sizes $\vec{\Delta}$,
as well as splitting the network into two branches processing the mask and the image separately, and only converging them in the bottom layers of the network.
The last option would allow precomputing the image-specific feature maps for the whole volume of interest, which could be done independently from the
single-object segmentation.

Qualitative analysis of segmentations produced by FFNs suggests at least two routes towards further increases of segmentation quality. One route involves
adding explicit information about neural ultrastructure (e.g., mitochondria) to the training procedure, which could reduce the number of cases where an organelle
is segmented as a separate fragment. The other route would be to address splits caused by extremely thin parts of neural processes during training
by, for example, increasing the frequency of training examples exhibiting such a problem or reweighting individual voxels based on a distance transform.

A major component of the FFN system that is currently not subject to explicit numerical optimization is the heuristic FoV movement procedure
used during training and inference. Instead of relying on a fixed step size and max probability direction, we envision a version of the FFN
that uses a learned policy and is thus able to dynamically adjust its movement pattern, as well as the scale at which its input data is processed.
Beyond allowing the whole system to be optimized end-to-end, this should also positively impact computational cost as the network could spend
less time on easy, unambiguous cases and concentrate the processing effort on ones that require more attention. Such a policy could be learned
either in a supervised or in a reinforcement learning setting.

All the above issues are the subject of current experiments, the results of which will be reported elsewhere.

\bibliographystyle{plainnat}
\bibliography{ffn}

\end{document}